\documentclass[conference]{IEEEtran}

\usepackage{amsmath,amssymb}
\usepackage{booktabs}
\usepackage{balance}
\usepackage{cite}
\usepackage{graphicx}
\usepackage{multirow}
\usepackage{url}

\newcommand{\kWp}{\mathrm{kW}_{\mathrm{p}}}
\newcommand{\best}[1]{\mathbf{#1}}

\title{Segmentation-Guided Direct Capacity Estimation of Rooftop Photovoltaics from Aerial Imagery}

\author{\IEEEauthorblockN{Guibin Chen\IEEEauthorrefmark{1}, Emma Fan\IEEEauthorrefmark{2}, and Yuzhang Lin\IEEEauthorrefmark{1}}
\IEEEauthorblockA{Department of Electrical and Computer Engineering, New York University\\
New York, USA\\\IEEEauthorrefmark{1}\{gc3280, yuzhang.lin\}@nyu.edu\\
\IEEEauthorrefmark{2}emma1q.fan@gmail.com}}

\begin{document}
\maketitle

\begin{abstract}
Distribution planners need installation-level inventories of rooftop photovoltaic (PV) capacity, but interconnection records may be incomplete or spatially aggregated. Conventional image-based estimators segment PV arrays, convert the mask to physical area, and multiply by an estimated power density, coupling capacity prediction to mask-derived area. We propose Direct Head, a segmentation-guided architecture that retains U-Net segmentation and soft-mask pooling while directly regressing nonnegative capacity from PV-region features and area descriptors. We evaluate three independent runs on 7,590 French rooftop installations under the original split and a department-based holdout. Relative to a parameter-matched area-density neural model, Direct Head reduces mean absolute error by 8.7\% and 12.5\%, and symmetric mean absolute percentage error by 14.0\% and 20.7\%, respectively. Comparisons with conventional estimators, controlled architecture variants, auxiliary-loss ablations, and representative cases show that segmentation provides useful spatial evidence without requiring capacity to be determined by an area-density product.
\end{abstract}


\section{Introduction}
Accurate rooftop photovoltaic (PV) inventories support distribution planning, hosting-capacity analysis, net-load forecasting, and the monitoring of distributed energy resources. The effects of rooftop PV on distribution feeders depend not only on aggregate installed capacity but also on the capacities and locations of individual installations: spatially concentrated PV can alter voltage profiles, cause reverse power flow, and reduce the remaining hosting capacity for additional distributed generation \cite{dubey2015hosting,wu2023hosting}. Installation-level location and nameplate capacity are therefore useful planning inputs, although they do not replace electrical feeder models.

Utilities typically obtain PV inventories from interconnection records, permitting systems, or customer programs. These sources can lag field deployment, omit older or unauthorized systems, and report locations at an aggregated spatial unit. Public and crowdsourced databases extend coverage but introduce geolocation, duplicate, and label-quality problems \cite{kasmi2023dataset,hu2022pitfalls}. High-resolution aerial imagery provides an independent and spatially resolved observation of equipment visible on rooftops.

PV mapping studies have progressed from engineered overhead-image classifiers and public polygon datasets \cite{malof2016automatic,bradbury2016dataset} to large-scale convolutional systems such as DeepSolar and SolarFinder \cite{yu2018deepsolar,li2020solarfinder}. Capacity estimation, however, is commonly treated as a two-stage physical proxy: panel extent is extracted and then converted using a constant, learned, or technology-dependent power density \cite{so2017capacity,tremenbert2023pypvroof}. Recent classification-aware approaches further distinguish PV types from aerial imagery \cite{wang2025classified}. These methods are interpretable, but the equality $C=A\rho$ becomes restrictive when a top-down mask misses tilted surfaces, occluded modules, or parts of an installation. Projected area also does not reveal module efficiency, inverter sizing, or registry noise.

For distribution-system applications, panel mapping and capacity inventory construction are related but distinct tasks. A binary footprint establishes where equipment is visible, whereas planning studies require an estimate of the rated power connected behind each service location. Consequently, an accurate mask can still yield a biased capacity estimate when projected area is not proportional to installed module area or when electrical ratings differ across visually similar arrays. This motivates an estimator that preserves spatial evidence from segmentation while allowing the final capacity mapping to depart from a fixed physical conversion.

This paper introduces Direct Head to study whether segmentation should guide capacity estimation without determining it algebraically. A U-Net \cite{ronneberger2015unet} predicts a PV soft mask, the mask pools bottleneck features, and a compact multilayer perceptron (MLP) predicts capacity directly. Segmentation and area supervision structure the learned representation, while the deployed output does not enforce $\widehat C=\widehat A\widehat\rho$. The design therefore preserves spatial supervision while allowing appearance features to compensate for an imperfect projected-area estimate.

Segmentation accuracy and capacity accuracy answer different questions. A segmentation mask represents the visible planimetric footprint of a PV array, whereas registry capacity reflects rated electrical output. Direct Head occupies the middle ground between hard physical factorization and unconstrained image regression: the mask remains an auditable intermediate product and suppresses irrelevant background, while the final mapping from visible evidence to capacity is learned from data.

\begin{figure*}[t]
  \centering
  \includegraphics[width=0.8\textwidth]{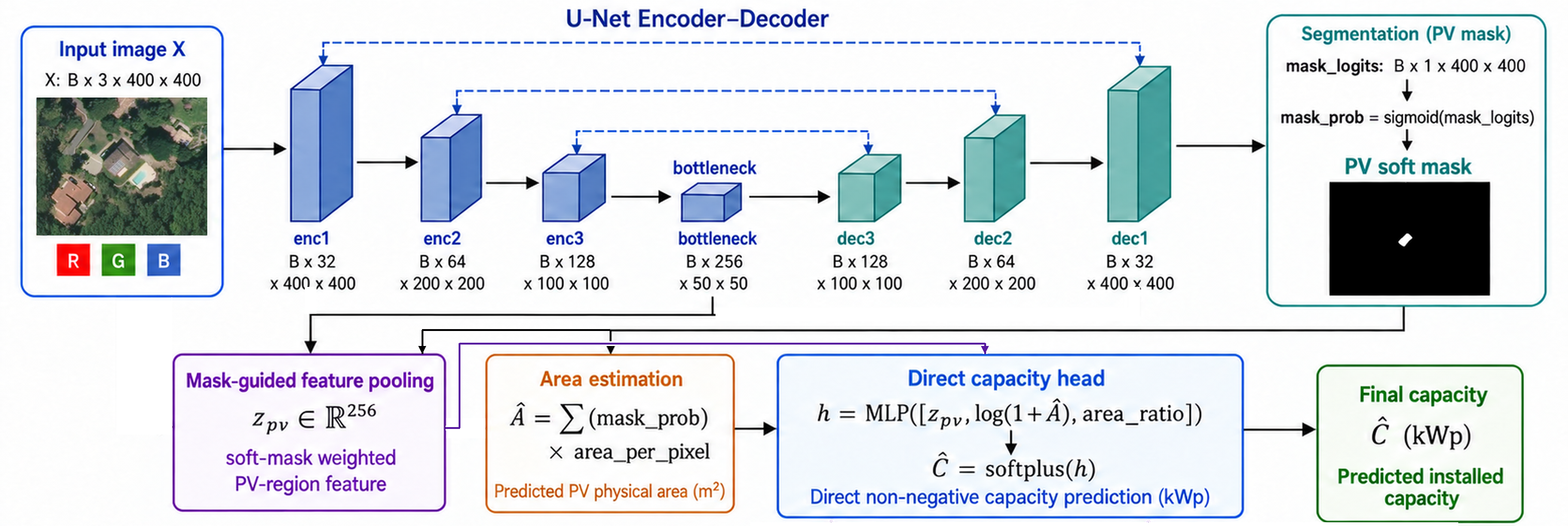}
  \caption{Direct Head inference architecture. A U-Net supplies a PV soft mask and bottleneck features. The mask guides feature pooling and area descriptors, while installed capacity is predicted directly. Training losses are defined in (\ref{eq:loss}).}
  \label{fig:architecture}
\end{figure*}

Installation-level capacity inventories complement, but do not replace, time-varying solar forecasting. Recent studies have addressed regional distributed PV ultra-short-term power forecasting under limited data availability \cite{wang2024domain} and short-term solar irradiance forecasting using a lightweight edge-computing framework \cite{wang2024irradiance}. These forecasting tasks concern time-varying quantities, whereas this work estimates static nameplate capacities from aerial imagery.

The main contributions are as follows: 1) A segmentation-guided direct capacity regressor that shares its backbone with the Area-density neural model and matches its parameter count to reporting precision, without enforcing an area--density product in the output. 2) A controlled comparison under the original split and a department-based holdout, using three independently trained seeds and a common model-selection rule. 3) An evaluation on a large real-world dataset against classical area-based regressors, neural architecture variants, and ablations of area and segmentation supervision, supplemented by representative cases relating segmentation quality to capacity error.

\section{Methodology}
\subsection{Segmentation-Guided Direct Head}
For a red-green-blue (RGB) crop $X\in\mathbb{R}^{3\times H\times W}$, a four-stage U-Net \cite{ronneberger2015unet} produces mask logits and a soft mask $P=\sigma(f_{\mathrm{seg}}(X))$. If $a_{\mathrm{px}}$ denotes the physical area represented by one resized pixel, the estimated visible PV area is
\begin{equation}
 \widehat A=a_{\mathrm{px}}\sum_{i,j}P_{ij}.
 \label{eq:area}
\end{equation}
The mask is resized to the bottleneck grid and used for normalized feature pooling,
\begin{equation}
 z_{\mathrm{pv}}=\frac{\sum_{i,j}\widetilde P_{ij}F_{ij}}
 {\sum_{i,j}\widetilde P_{ij}+\epsilon},
 \label{eq:pool}
\end{equation}
where $F\in\mathbb{R}^{256\times h\times w}$ is the bottleneck tensor. The Direct Head receives $z_{\mathrm{pv}}$, $\log(1+\widehat A)$, and the mean mask probability $r_A=(HW)^{-1}\sum_{i,j}P_{ij}$. It outputs estimated capacity
\begin{equation}
 \widehat C=\operatorname{softplus}\!\left(f_{\mathrm{cap}}
 ([z_{\mathrm{pv}},\log(1+\widehat A),r_A])\right).
 \label{eq:head}
\end{equation}
Area remains an informative feature but cannot mechanically scale the result. The logarithm compresses the long-tailed area range, and $r_A$ provides a scale-normalized occupancy cue. Soft-mask pooling suppresses background responses without imposing a hard threshold. The model contains 1.99 million trainable parameters, matching the Area-density neural model to reporting precision.

The complete computation graph is shown in Fig.~\ref{fig:architecture}. The segmentation branch supplies both an auditable spatial output and the weighting map used to form the PV-region descriptor, while the capacity branch produces the deployed scalar estimate. Annotated PV area provides an auxiliary supervision signal for the scale descriptor during training.

Direct regression retains the physical cues of an area-based estimator: (\ref{eq:head}) exposes the head to visible extent, site occupancy, and mask-conditioned appearance, while segmentation anchors the encoder spatially. It combines these cues through a learned nonlinear map. Unlike the Area-density neural model, Direct Head does not require capacity to be computed as the product of predicted area and power density. Although mask errors can still affect the pooled features and area descriptors, the learned regression head can adjust their influence rather than propagating them through a compulsory product. The softplus output enforces nonnegativity without imposing a fixed upper bound on predicted capacity.

The information paths in Fig.~\ref{fig:architecture} are deliberately asymmetric. Segmentation determines where visual evidence is pooled, but it does not algebraically determine the deployed output. The capacity objective can refine the shared representation without removing pixel-level supervision, and the mask, area estimate, and final prediction remain separately inspectable.

Pooling at the bottleneck balances localization and semantic abstraction. The normalized average in (\ref{eq:pool}) prevents feature magnitude from increasing merely because more pixels are activated, while $\log(1+\widehat A)$ and $r_A$ supply explicit scale information. Because the soft mask is retained rather than thresholded inside the head, boundary uncertainty remains available to the model and capacity gradients can refine the shared representation. The deployed output therefore remains accompanied by an inspectable mask and area descriptor.

\subsection{Training Objective}
Capacity and area are trained in logarithmic space with Huber losses to reduce the influence of rare large installations. The segmentation loss $\mathcal L_{\mathrm{seg}}$ combines binary cross-entropy and Dice loss to provide pixel-level supervision. With annotated mask area $A^{*}$, the training objective is
\begin{equation}
 \mathcal L=\mathcal L_{\mathrm{seg}}+\mathcal L_{\mathrm{cap}}
 +0.2\mathcal L_{\mathrm{area}}.
 \label{eq:loss}
\end{equation}
Here, $\mathcal L_{\mathrm{cap}}$ compares $\log(1+\widehat C)$ and $\log(1+C)$, while $\mathcal L_{\mathrm{area}}$ compares $\log(1+\widehat A)$ and $\log(1+A^{*})$. The capacity term optimizes the quantity used in deployment, while the area term encourages a physically meaningful scale descriptor. The coefficient 0.2 keeps area supervision auxiliary to the capacity objective.

All three loss terms are used during training. At inference, only the image and its physical scale are required; the network directly predicts capacity while also producing an inspectable soft mask and area estimate.

\section{Experimental Setup}
\subsection{Data and Evaluation Protocols}
We use the aerial-imagery subset of the PV Database for Deep Learning (BDAPPV), which provides RGB images, crowd-annotated segmentation masks, installation metadata, and nameplate capacities \cite{kasmi2023dataset}. Its 7,590 French rooftop PV image chips were extracted from geometrically corrected aerial imagery provided by the French National Institute of Geographic and Forest Information (IGN), with a ground sampling distance of 0.20~m/pixel. Capacity ranges from 0.12 to 258.3 kilowatts-peak ($\kWp$), with a median of 2.96~$\kWp$, yielding a strongly right-skewed regression problem. Formal experiments retain the native $400\times400$ image resolution.

The original protocol contains 6,072/759/759 training/validation/test samples, respectively, for a total of 7,590 installations. To complement this evaluation, we construct a department-based protocol using French administrative department identifiers. We exclude 102 records without valid department codes because they cannot be reliably assigned to an administrative department, leaving 7,488 installations. The resulting split contains 5,759/858/871 training/validation/test samples.

Unlike a sample-level split, the administrative protocol assigns entire departments to mutually exclusive training, validation, and test partitions. All installations from the same department therefore remain in one partition. This design evaluates model performance in departments not represented during training and reduces geographic overlap between the training and evaluation regions. It is particularly relevant to installation-level capacity mapping, where acquisition conditions, rooftop characteristics, and the distribution of PV system sizes may differ among geographic areas.

The original split is retained to preserve comparability with the established dataset partition. The department-based protocol provides a complementary evaluation of within-dataset regional transfer. By testing both protocols with the same modeling and evaluation procedure, we examine whether the relative ranking of Direct Head and the baseline methods is robust to the choice of data partition rather than relying on a single split.

\subsection{Baselines and Implementation}
The Area-density neural model uses the same U-Net backbone, pooled features, MLP dimensions, and parameter count as Direct Head, but enforces $\widehat C=\widehat A\widehat\rho$ with bounded $\widehat\rho$. Motivated by prior aerial-image capacity estimation methods \cite{so2017capacity,tremenbert2023pypvroof}, we also consider three conventional capacity-estimation baselines: a constant power-density estimator fitted on the training set, an area-based linear regression model, and a random forest with 200 trees. The first two methods use predicted PV area, while the random forest additionally incorporates mask-conditioned color statistics. Regressors using annotated test masks are treated as oracle-input diagnostics and excluded from the deployable model comparison.

To evaluate individual architectural components of the proposed Direct Head, we further conduct two controlled ablation experiments, reported in Table~\ref{tab:architecture}. The global-pooling (GP) variant replaces the mask-guided feature pooling in Direct Head with spatial average pooling. The no-capacity-feedback (NCF) variant blocks capacity-loss gradients from propagating through part of the shared segmentation pathway.

All neural models use 60 epochs, batch size 8, AdamW with learning rate $3\times10^{-4}$ and weight decay $10^{-4}$, dropout 0.1, and strict deterministic execution on a graphics processing unit (GPU). Checkpoint selection uses the same frozen validation criterion for every neural configuration. Every formal configuration is independently trained with seeds 0, 1, and 2 on NVIDIA A100 GPUs and evaluated once on the test split.

Both manifests are frozen before training, and each run records its configuration, source revision, validation history, checkpoint, and row-level predictions. Neural comparisons share preprocessing, data order, optimizer, epoch budget, and validation rule. Classical regressors are fitted per seed from the corresponding segmentation predictions; test labels are not used for model selection.

The baseline set separates fixed physical conversion, learned area-only conversion, and nonlinear area--appearance regression. The Area-density neural model additionally holds the segmentation backbone, pooled descriptor, and head size fixed while changing the output equation to $\widehat A\widehat\rho$. It therefore provides the most direct test of the proposed output parameterization.

We evaluate capacity estimation using mean absolute error (MAE), reported in $\kWp$, and symmetric mean absolute percentage error (sMAPE). Segmentation performance is evaluated using intersection over union (IoU). The relative-error metric is
\begin{equation}
 \mathrm{sMAPE}=\frac{100}{N}\sum_n
 \frac{2|C_n-\widehat C_n|}{|C_n|+|\widehat C_n|}.
\end{equation}
MAE is the primary capacity metric for the predominantly residential systems, sMAPE describes relative error across the capacity range, and IoU measures agreement between the thresholded predicted mask and the annotated PV mask. Results are the mean and sample standard deviation over three seeds.

\section{Results and Discussion}
\subsection{Comparison with Capacity-Estimation Baselines}
\begin{table*}[t]
\centering
\caption{Formal test performance (mean $\pm$ sample standard deviation over three seeds).}
\label{tab:main}
\setlength{\tabcolsep}{6.0pt}
\begin{tabular}{llccc}
\toprule
Protocol & Method & MAE ($\kWp$) & sMAPE (\%) & IoU \\
\midrule
\multirow{5}{*}{Original}
 & Area $\times$ constant density & $1.821\pm0.017$ & $31.91\pm1.43$ & $0.744\pm0.004$ \\
 & Area linear regression & $1.677\pm0.029$ & $27.28\pm1.31$ & $0.744\pm0.004$ \\
 & Area + color random forest & $1.253\pm0.015$ & $19.61\pm0.35$ & $0.744\pm0.004$ \\
 & Area-density neural model & $1.345\pm0.073$ & $20.79\pm1.90$ & $0.718\pm0.020$ \\
 & \textbf{Direct Head (proposed)} & $\best{1.228\pm0.028}$ & $\best{17.87\pm0.24}$ & $0.724\pm0.010$ \\
\midrule
\multirow{5}{*}{Administrative}
 & Area $\times$ constant density & $1.607\pm0.059$ & $29.00\pm1.50$ & $0.763\pm0.009$ \\
 & Area linear regression & $1.549\pm0.070$ & $26.80\pm2.00$ & $0.763\pm0.009$ \\
 & Area + color random forest & $1.107\pm0.016$ & $17.07\pm0.15$ & $0.763\pm0.009$ \\
 & Area-density neural model & $1.182\pm0.023$ & $19.46\pm0.60$ & $0.739\pm0.008$ \\
 & \textbf{Direct Head (proposed)} & $\best{1.034\pm0.018}$ & $\best{15.44\pm0.29}$ & $0.746\pm0.001$ \\
\bottomrule
\end{tabular}
\end{table*}

Table~\ref{tab:main} shows that the final Direct Head obtains the lowest MAE and sMAPE under both protocols. Relative to the Area-density neural model, it reduces MAE by 8.7\% on the original split and 12.5\% on the administrative holdout; sMAPE decreases by 14.0\% and 20.7\%, respectively. The area-color random forest remains competitive, but Direct Head improves its MAE by 2.0\% and 6.6\% and its sMAPE by 8.9\% and 9.5\%. The shared classical mask source attains higher IoU than Direct Head, reinforcing that spatial overlap and capacity accuracy are related but distinct objectives.

The consistent ranking across the two protocols is important because the department-based holdout separates geographic regions between training and evaluation while changing the composition of the test population. Direct Head therefore does not obtain its advantage from only one partitioning rule. The proposed model improves on that reference by learning a joint representation rather than by discarding area information.

Absolute errors are lower under the administrative protocol for all leading methods. This does not imply that regional transfer is intrinsically easier: the protocol changes the retained records and test capacity distribution as well as the geographic grouping. It is appropriate for within-protocol method ranking, while direct MAE comparison between protocols conflates partition composition and generalization difficulty.

MAE weights installations linearly, whereas sMAPE normalizes each target--prediction pair. Direct Head improves both under both protocols, so its advantage is not explained solely by changing prediction scale. These results are consistent with the hypothesis that mask-conditioned appearance features provide useful capacity information beyond the estimated PV area alone. The IoU columns provide the complementary observation that the best spatial overlap does not necessarily yield the best capacity estimate.

\subsection{Controlled Architecture Analysis}
\begin{table*}[t]
\centering
\caption{Controlled neural-model analysis (mean $\pm$ sample standard deviation over three seeds).}
\label{tab:architecture}
\setlength{\tabcolsep}{3.2pt}
\footnotesize
\begin{tabular}{lcccccc}
\toprule
& \multicolumn{3}{c}{Original} & \multicolumn{3}{c}{Administrative} \\
\cmidrule(lr){2-4}\cmidrule(lr){5-7}
Variant & MAE & sMAPE & IoU & MAE & sMAPE & IoU \\
\midrule
Area-density neural model & $1.345\pm0.073$ & $20.79\pm1.90$ & $0.718\pm0.020$ & $1.182\pm0.023$ & $19.46\pm0.60$ & $0.739\pm0.008$ \\
Global-pooling (GP) & $1.283\pm0.037$ & $19.49\pm0.66$ & $0.742\pm0.004$ & $1.201\pm0.065$ & $19.38\pm1.80$ & $0.748\pm0.019$ \\
No-capacity-feedback (NCF) & $1.322\pm0.066$ & $20.89\pm0.89$ & $\best{0.744\pm0.004}$ & $1.224\pm0.034$ & $20.00\pm0.65$ & $\best{0.763\pm0.009}$ \\
\textbf{Direct Head} & $\best{1.228\pm0.028}$ & $\best{17.87\pm0.24}$ & $0.724\pm0.010$ & $\best{1.034\pm0.018}$ & $\best{15.44\pm0.29}$ & $0.746\pm0.001$ \\
\bottomrule
\end{tabular}
\end{table*}

Table~\ref{tab:architecture} compares the proposed Direct Head with the Area-density neural model and two controlled architecture variants. Direct Head achieves the lowest MAE and sMAPE among the tested neural configurations under both protocols. Relative to the Area-density neural model, its MAE decreases from 1.345 to 1.228~$\kWp$ on the original split and from 1.182 to 1.034~$\kWp$ on the administrative holdout. The corresponding sMAPE values decrease from 20.79\% to 17.87\% and from 19.46\% to 15.44\%, respectively.

The other architectural variants exhibit less consistent behavior. Global pooling improves MAE over the Area-density neural model on the original split but performs slightly worse on the administrative holdout. Blocking capacity feedback also has protocol-dependent effects and does not outperform Direct Head. 
These results demonstrate that Direct Head consistently achieves the best MAE and sMAPE among all evaluated neural architectures under both evaluation protocols. Its superior performance highlights the effectiveness of segmentation-guided direct capacity regression for rooftop PV capacity estimation.

\subsection{Auxiliary Supervision and Qualitative Behavior}
\begin{figure}[t]
  \centering
  \includegraphics[width=0.85\columnwidth]{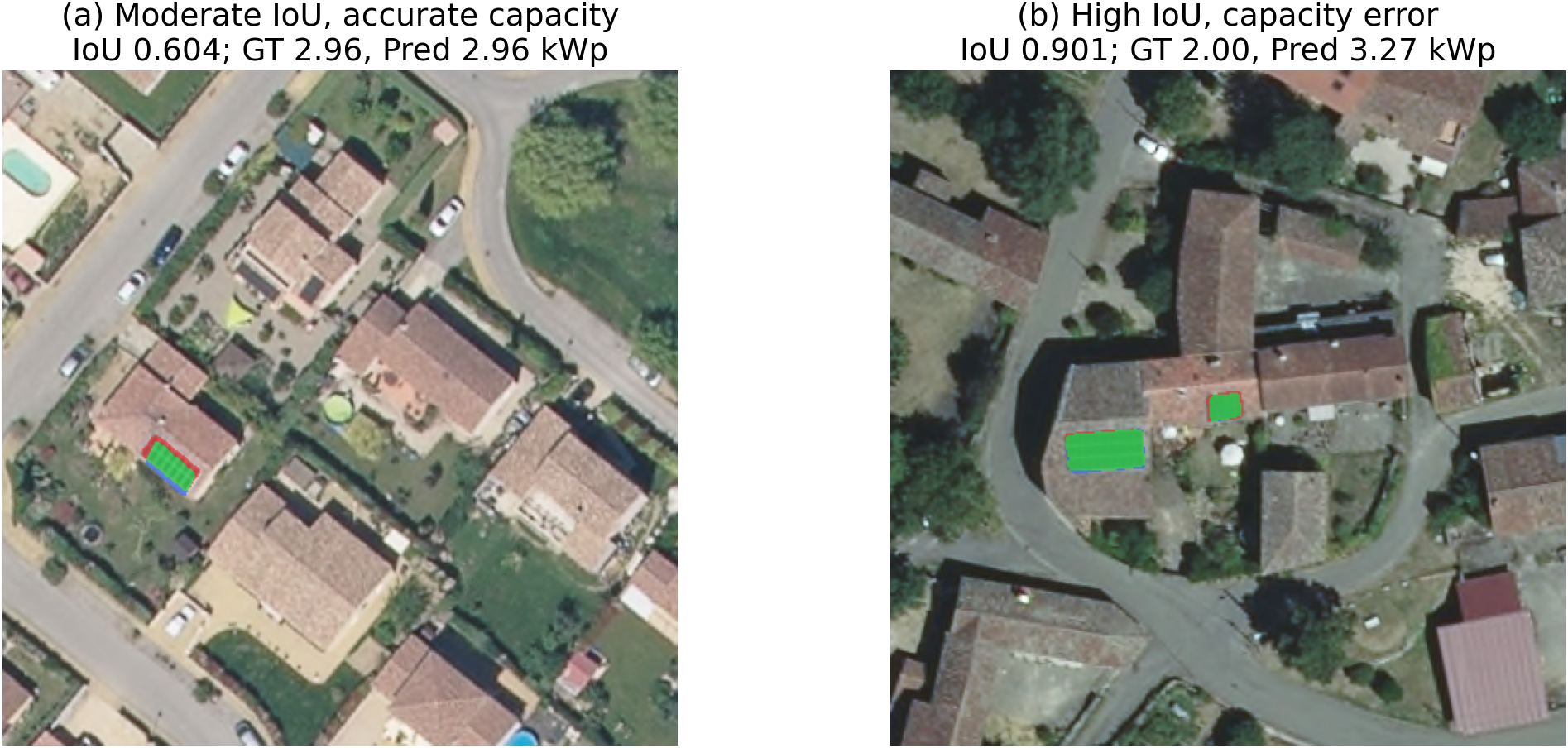}
  \caption{Contrasting Direct Head test cases: low IoU with accurate capacity and high IoU with appreciable capacity error. Green, red, and blue denote true-positive, false-positive, and false-negative mask pixels.}
  \label{fig:qualitative}
\end{figure}
\vspace{-5pt}
\begin{table}[t]
\centering
\caption{Ablation study of the proposed Direct Head on validation set.}
\label{tab:auxiliary}
\setlength{\tabcolsep}{2.5pt}
\footnotesize
\begin{tabular}{lccc}
\toprule
Configuration & MAE & sMAPE & IoU \\
\midrule
\textbf{Full objective} & $\best{1.577}$ & 16.854 & $\best{0.733}$ \\
Without area loss & 1.615 & $\best{16.839}$ & 0.728 \\
Without segmentation loss & 1.736 & 18.594 & 0.085 \\
\bottomrule
\end{tabular}
\end{table}

Table~\ref{tab:auxiliary} reports ablations of the area and segmentation losses in the final Direct Head objective in (\ref{eq:loss}). Removing area supervision raises validation MAE from 1.577 to 1.615~$\kWp$ and slightly reduces IoU from 0.733 to 0.728, while sMAPE changes marginally from 16.854\% to 16.839\%. Removing segmentation supervision raises MAE to 1.736~$\kWp$ and sharply reduces IoU to 0.085. These observations support the role of pixel-level segmentation supervision and suggest a more modest contribution from area supervision in the reduced setting. 

Fig.~\ref{fig:qualitative} illustrates why segmentation should guide, but not algebraically determine, capacity estimation. A low-IoU case retains an accurate capacity estimate, whereas a high-IoU case has appreciable capacity error. The latter behavior is consistent with differences in module efficiency, orientation, inverter sizing, and possible registry-label uncertainty that cannot be determined reliably from the visible projected footprint alone.

An exploratory three-seed analysis using an earlier model configuration showed that a run with lower segmentation IoU (0.629 versus 0.687) also achieved lower capacity MAE (1.608 versus 1.629~$\kWp$). This observation further illustrates that segmentation accuracy and capacity estimation accuracy do not necessarily improve together.

These cases are illustrative rather than a substitute for aggregate evaluation. They show that spatial overlap and installed-capacity error need not move together: an incomplete mask can still retain informative PV-region features, while an accurate footprint does not reveal every electrical attribute associated with nameplate capacity. In practical inventory reconciliation, the predicted mask, derived area, and direct capacity estimate can be inspected jointly to prioritize installations for manual review.

\section{Conclusion}
This paper separates the value of PV segmentation from the constraint of area-based capacity computation. A soft mask guides feature pooling, while direct nonnegative regression estimates installation nameplate capacity. Across two evaluation protocols and three independently trained seeds on a large real-world dataset, Direct Head achieves lower MAE and sMAPE than the parameter-matched Area-density neural model and an area--color random forest. Controlled comparisons support the effectiveness of the tested direct-output configuration, while representative cases illustrate that accurate segmentation alone does not guarantee accurate capacity estimation. The approach preserves an auditable spatial output without forcing projected area to determine electrical capacity. Future work should validate across image providers and acquisition years, quantify predictive uncertainty, and examine how installation-level estimation errors affect feeder-level planning studies.

\balance


\begin{thebibliography}{14}
\bibitem{dubey2015hosting}
A. Dubey, S. Santoso, and A. Maitra, ``Understanding photovoltaic hosting capacity of distribution circuits,'' in \emph{Proc. 2015 IEEE Power Energy Soc. Gen. Meeting (PESGM)}, 2015, pp. 1--5, doi: 10.1109/PESGM.2015.7286510.

\bibitem{wu2023hosting}
J. Wu, J. Yuan, Y. Weng, and R. Ayyanar, ``Learn dynamic hosting capacity based on voltage sensitivity analysis,'' in \emph{Proc. 2023 IEEE Power Energy Soc. Gen. Meeting (PESGM)}, 2023, pp. 1--5, doi: 10.1109/PESGM52003.2023.10252543.

\bibitem{kasmi2023dataset}
G. Kasmi \emph{et al.}, ``A crowdsourced dataset of aerial images with annotated solar photovoltaic arrays and installation metadata,'' \emph{Sci. Data}, vol. 10, art. no. 59, Jan. 2023, doi: 10.1038/s41597-023-01951-4.

\bibitem{hu2022pitfalls}
W. Hu \emph{et al.}, ``What you get is not always what you see---pitfalls in solar array assessment using overhead imagery,'' \emph{Appl. Energy}, vol. 327, art. no. 120143, Dec. 2022, doi: 10.1016/j.apenergy.2022.120143.

\bibitem{malof2016automatic}
J. M. Malof, K. Bradbury, L. M. Collins, and R. G. Newell, ``Automatic detection of solar photovoltaic arrays in high resolution aerial imagery,'' \emph{Appl. Energy}, vol. 183, pp. 229--240, Dec. 2016, doi: 10.1016/j.apenergy.2016.08.191.

\bibitem{bradbury2016dataset}
K. Bradbury \emph{et al.}, ``Distributed solar photovoltaic array location and extent dataset for remote sensing object identification,'' \emph{Sci. Data}, vol. 3, art. no. 160106, Dec. 2016, doi: 10.1038/sdata.2016.106.

\bibitem{yu2018deepsolar}
J. Yu, Z. Wang, A. Majumdar, and R. Rajagopal, ``DeepSolar: A machine learning framework to efficiently construct a solar deployment database in the United States,'' \emph{Joule}, vol. 2, no. 12, pp. 2605--2617, Dec. 2018, doi: 10.1016/j.joule.2018.11.021.

\bibitem{li2020solarfinder}
Q. Li, Y. Feng, Y. Leng, and D. Chen, ``SolarFinder: Automatic detection of solar photovoltaic arrays,'' in \emph{Proc. 19th ACM/IEEE Int. Conf. Inf. Process. Sensor Netw. (IPSN)}, Sydney, NSW, Australia, 2020, pp. 193--204, doi: 10.1109/IPSN48710.2020.00024.

\bibitem{so2017capacity}
B. So \emph{et al.}, ``Estimating the electricity generation capacity of solar photovoltaic arrays using only color aerial imagery,'' in \emph{Proc. 2017 IEEE Int. Geosci. Remote Sens. Symp. (IGARSS)}, Fort Worth, TX, USA, 2017, pp. 1603--1606, doi: 10.1109/IGARSS.2017.8127279.

\bibitem{tremenbert2023pypvroof}
Y. Tr{\'e}menbert, G. Kasmi, L. Dubus, Y.-M. Saint-Drenan, and P. Blanc, ``PyPVRoof: A Python package for extracting the characteristics of rooftop PV installations using remote sensing data,'' arXiv:2309.07143, Sep. 2023, doi: 10.48550/arXiv.2309.07143.

\bibitem{wang2025classified}
K. Li, M. Gong, Z. Li, and C. Huang, ``Classified identification and estimation of behind-the-meter distributed photovoltaic panels using high-resolution aerial imagery,'' \emph{IEEE Trans. Ind. Appl.}, vol. 61, no. 1, pp. 1604--1612, Jan./Feb. 2025, doi: 10.1109/TIA.2024.3384139.

\bibitem{ronneberger2015unet}
O. Ronneberger, P. Fischer, and T. Brox, ``U-Net: Convolutional networks for biomedical image segmentation,'' in \emph{Proc. 18th Int. Conf. Med. Image Comput. Comput.-Assist. Interv. (MICCAI)}, Munich, Germany, 2015, pp. 234--241, doi: 10.1007/978-3-319-24574-4\_28.

\bibitem{wang2024domain}
Y. Wang, H. Dai, Z. Zhen, F. Wang, J. Wang, and L. Lu, ``Domain adversarial graph neural network based regional distributed PV ultra-short-term power forecasting considering data-scarce scenarios,'' in \emph{Proc. 2024 IEEE Power Energy Soc. Gen. Meeting (PESGM)}, 2024, pp. 1--5, doi: 10.1109/PESGM51994.2024.10689089.

\bibitem{wang2024irradiance}
S. Wang, W. Zhang, Y. Sun, V. Archana, and D. Srinivasan, ``Short-term forecasting of solar irradiance: A lightweight edge computing framework,'' in \emph{Proc. 2024 IEEE Power Energy Soc. Gen. Meeting (PESGM)}, 2024, pp. 1--5, doi: 10.1109/PESGM51994.2024.10688718.
\end{thebibliography}
\end{document}